\documentclass[letterpaper, 10 pt, conference]{ieeeconf}  

\IEEEoverridecommandlockouts                              

\usepackage{graphicx}
\usepackage{makecell}
\usepackage[T1]{fontenc}          
\usepackage{textcomp}             
\DeclareGraphicsExtensions{.pdf,.eps,.jpg,.png}
\usepackage{caption}
\usepackage{float} 
\usepackage{adjustbox}
\usepackage{subcaption}
\usepackage{cite}
\usepackage{multirow}
\usepackage{tabularx} 
\makeatletter
\let\NAT@parse\undefined
\makeatother
\usepackage[colorlinks=true,
linkcolor=blue,
anchorcolor=blue,
citecolor=blue
]{hyperref}  

\def\eqref#1{\textcolor{blue}{(\ref{#1})}}
\usepackage[all]{hypcap}
\usepackage{booktabs}
\usepackage{pdflscape}
\usepackage{array}
\usepackage{amsmath,amssymb}
\usepackage{siunitx}
\usepackage{algorithm}
\usepackage{algpseudocode}
\newcommand{\bea}{\begin{equation} \begin{aligned}}
\newcommand{\eea}{\end{aligned}\end{equation}}
\def\dif{\mathop{}\hphantom{\mskip-\thinmuskip}\mathrm{d}}%
\let\daccent\d
\let\d\relax
\newcommand\dd{\ifmmode\dif\else\expandafter\daccent\fi}

\title{\LARGE \bf
USIM and U0: A Vision-Language-Action Dataset and Model for General Underwater Robots
}

\author{Junwen~Gu\textsuperscript{1,3$\dagger$}, Zhiheng~Wu\textsuperscript{2$\dagger$}, Pengxuan~Si\textsuperscript{3}, Shuang~Qiu\textsuperscript{1,3}, Zhentao~Zhang\textsuperscript{1,3}, Yukai~Feng\textsuperscript{1,3}, \\Luoyang~Sun\textsuperscript{1,3}, Laien~Luo\textsuperscript{1,3}, Lianyi~Yu\textsuperscript{1,3}, Jian~Wang\textsuperscript{1,3}, and Zhengxing~Wu\textsuperscript{1,3*}
\thanks{\emph{Junwen~Gu and Zhiheng~Wu contributed equally to this work. Corresponding author: Zhengxing~Wu.}}
\thanks{\textsuperscript{1}The Key Laboratory of Cognition and Decision Intelligence for Complex Systems, Institute of Automation, Chinese Academy of Sciences, Beijing 100190, China (e-mail: gujunwen2022@ia.ac.cn jianwang@ia.ac.cn; zhengxing.wu@ia.ac.cn).}%
\thanks{\textsuperscript{2}Baidu Inc., Beijing 100085, China (e-mail: wzh404.ai@gmail.com).}%
\thanks{\textsuperscript{3}The School of Artificial Intelligence, University of Chinese Academy of Sciences, Beijing 100049, China.}%
\thanks{
    \texttt{\href{https://vincentgu2000.github.io/u0project}{https://vincentgu2000.github.io/u0project}}
}%
}

\begin{document}

\maketitle
\thispagestyle{empty}
\pagestyle{empty}

\begin{abstract}

Underwater environments pose unique challenges for robotic navigation and manipulation. While existing research has primarily focused on task-specific methods, studies on general-purpose intelligence for multi-task execution remain scarce. To address this gap, we propose a unified framework for general-purpose underwater robots that integrates perception and action driven by language instructions. First, we develop a data synthesis pipeline to construct USIM, a simulation-based dataset which comprises over 905K frames from 2275 trajectories, totaling approximately 25 hours of BlueROV2 interactions. Furthermore, we propose U0, a vision-language-action (VLA) model capable of executing various tasks from obstacle-avoidance navigation to three-dimensional mobile manipulation. The model features a convolution-attention-based perception (CAP) module, which incorporates target pose estimation as an auxiliary task to explicitly bolster the model's spatial awareness. For evaluation, we establish a systematic assessment framework and an automated pipeline encompassing both offline metrics and online task execution. Experimental results demonstrate that the USIM dataset significantly empowers existing VLA models to adapt to underwater scenarios. Notably, our U0 model achieves state-of-the-art performance: it reduces the offline mean action prediction error to 0.0359 and achieves an overall online success rate of 43.1\%, marking a 5.5\% improvement over existing competitive baselines (below 37.6\%), with navigation tasks reaching as high as 87.5\%. These results validate the feasibility of general-purpose intelligence in underwater robotics, providing a foundation for scalable dataset synthesis and aquatic embodied agents.

\end{abstract}

\begin{table*}
    \centering
    \caption{Comparison of Typical Underwater Datasets and USIM Dataset}
    \label{dataset_comparison}
    \small
    \renewcommand{\arraystretch}{1.3}
    \setlength{\tabcolsep}{4pt}
    \resizebox{\textwidth}{!}{
    \begin{tabular}{l c r r r l l l c}
        \toprule
        \textbf{Dataset} & \textbf{Type} & $\mathbf{N_{traj}}$ & $\mathbf{N_{fr.}}$ & \textbf{Dur.} & \textbf{Primary Tasks} & \textbf{Modality} & \textbf{Labels} & \textbf{Lang.} \\
        \midrule
        Aqualoc~\cite{ferrera_aqualoc_2019}  & Real & 17    & --    & 1.7~h & Visual SLAM, VIO & Monochromatic, IMU, Depth & Est. pose & \texttimes \\
        MIMIR-UW~\cite{alvarez-tunon_mimir-uw_2023} & Sim  & 11    & 31K   & 0.4~h & SLAM, Seg., Depth Est. & RGBD, IMU & Pose, Sem. masks & \texttimes \\
        SubPipe~\cite{alvarez-tunon_subpipe_2024}  & Real & 5     & --    & 0.7~h & SLAM, Seg., Det. & RGB, SSS, IMU, DVL, Depth & Est. pose, Sem. masks & \texttimes \\
        MOUD~\cite{chu_multimodal_2025}     & Real & --    & 18K   & --   & Rec., Seg., Det. & RGB, LiDAR & Instance masks & \texttimes \\
        \midrule
        USIM (Ours) & Sim & 2275 & 905K & 25~h & Navigation, Manipulation & RGB, IMU, DVL, Depth & Pose, Action & \checkmark \\
        \bottomrule
    \end{tabular}
    }
    \vspace{1pt}
    \begin{flushleft}
        \scriptsize \textit{Note:} $\mathbf{N_{traj}}$, number of trajectories; $\mathbf{N_{fr.}}$, number of frames; Dur., duration; Lang., language; VIO, visual--inertial odometry; SSS, side scan sonar; Seg., segmentation; Est., estimation; Det., detection; Rec., reconstruction; Sem., semantic.
    \end{flushleft}
\end{table*}

\section{Introduction}

The underwater environment poses growing demands, encompassing diverse applications such as marine ecological surveys, resource exploitation, inspection of pipelines, and underwater salvage~\cite{liu_bioinspired_2025, katzschmann_exploration_2018}. However, in-situ human diving operations are far more challenging and hazardous than terrestrial or aerial tasks, which necessitating the use of underwater robots to execute these missions. Nevertheless, the development of these systems is hindered by distinctive environmental constraints, including complex hydrodynamics, limited visibility, and restricted communication. Enhancing the autonomy and intelligence of underwater robots is crucial to overcoming these challenges, as it empowers them to execute complex missions with minimal human intervention.

Existing studies have developed task-specialized frameworks for underwater teleoperation~\cite{phung_shared_2024}, navigation~\cite{hu_tightly-coupled_2022}, and manipulation~\cite{palmer_angler_2024}. Despite their success, the lack of a unified perspective results in fragmented data islands and complicates the integration of multi-task capabilities. Regarding the development of general embodied intelligence, prior works have explored reinforcement learning based approaches to achieve multi-task capability~\cite{bai_picor_2023}. In recent years, vision-language-action (VLA) models based on large-scale pre-training have catalyzed significant breakthroughs. By offering high autonomy and generalization, VLA models are inherently suited for complex underwater missions that demand multi-task capabilities. However, current VLA applications are primarily concentrated in robotic arm manipulation and quadrupedal navigation~\cite{nvidia_gr00t_2025, intelligence__05_2025, kim_openvla_2024, tong_quart-online_2025}. Since these methods are predominantly trained on terrestrial data biased toward two-dimensional (2D) planar motion, they are difficult to transfer directly to underwater robots with distinct configurations and three-dimensional (3D) degrees of freedom. Moreover, the vast discrepancies between terrestrial and aquatic scenes regarding visual signals pose significant challenges to a model's perception~\cite{xu_nautilus_2025}. Furthermore, data collection in real underwater environments is prohibitively costly; for instance, even experienced operators require extensive practice to achieve proficiency in teleoperated grasping tasks~\cite{liu_self-improving_2025}. Overall, the scarcity of unified frameworks and datasets stands in stark contrast to the urgent demand for higher autonomy and multi-task intelligence in underwater robotics.

To address these challenges, we establish a unified framework for general-purpose underwater robots, comprising the multi-task VLA dataset USIM and the corresponding VLA model U0, which collectively enable sophisticated underwater multi-task autonomy. Specifically, USIM is a simulation-based VLA dataset featuring both perception and control traces for the BlueROV2 platform. Developed via an automated data synthesis pipeline, USIM consists of 905K frames spanning 2275 trajectories and approximately 25 hours of robot-environment interactions. Covering 20 tasks across 9 scenarios ranging from navigation to mobile manipulation, it offers a rapid and cost-effective alternative to real-world data collection. By employing a stochastic exploratory strategy, USIM incorporates sub-optimal states and recovery behaviors, ensuring broad coverage of the state space--a critical factor for learning robust policies in dynamic aquatic environments. Building upon this data foundation, we propose U0, a 3B-parameter VLA model capable of simultaneously handling navigation and manipulation tasks. It utilizes a unified action space that directly operates low-level actuators, effectively addressing the unique embodiment of underwater robots that combines thruster locomotion with manipulator control. To tackle the challenges of expansive environments and sparse targets, a convolution-attention-based perception (CAP) module is designed to predict the poses of navigation points or grasping targets. This auxiliary training task explicitly bolsters the model's spatial awareness by introducing geometric constraints. To rigorously verify our approach, we establish a systematic assessment framework and an automated pipeline encompassing both offline metrics and online closed-loop execution. Extensive evaluations demonstrate that fine-tuning on the USIM dataset significantly reduces the action prediction error (APE) across all models, with U0 achieving the lowest APE of 0.0359. Furthermore, online results show that our framework achieves a state-of-the-art (SOTA) 43.1\% multi-task success rate, marking a 5.5\% absolute improvement over competitive baselines. Notably, U0 reaches an 87.5\% success rate in navigation tasks and 30.0\% in challenging 3D mobile manipulation—a domain where existing VLA models fail to perform (0\% success) without fine-tuning.

Overall, we present a unified framework that integrates a dataset with a VLA model tailored for diverse underwater missions. To the best of our knowledge, this represents the first effort to jointly address multi-task underwater perception and action guided by natural language instructions. The main contributions of this paper are summarized as follows:
\begin{itemize}
   \item A scalable, automated data synthesis pipeline and the USIM dataset, which significantly enhances model adaptability and policy robustness in dynamic aquatic environments.

   \item A general-purpose underwater VLA model, U0, featuring enhanced spatial awareness for precise perception and execution across diverse navigation and manipulation missions.
   
   \item A systematic assessment framework and a unified multi-task autonomy pipeline, setting a new benchmark for underwater robotics and demonstrating the feasibility of general-purpose underwater intelligence.
\end{itemize}

The remainder of this paper is organized as follows. Section~\ref{sec:related_work} reviews related work. Section~\ref{sec:method} details the proposed methodology. Section~\ref{sec:exp} presents experimental results and ablation studies. Section~\ref{sec:conclusion} concludes the paper.

\begin{figure*}
   \centering
   \includegraphics[width=\textwidth]{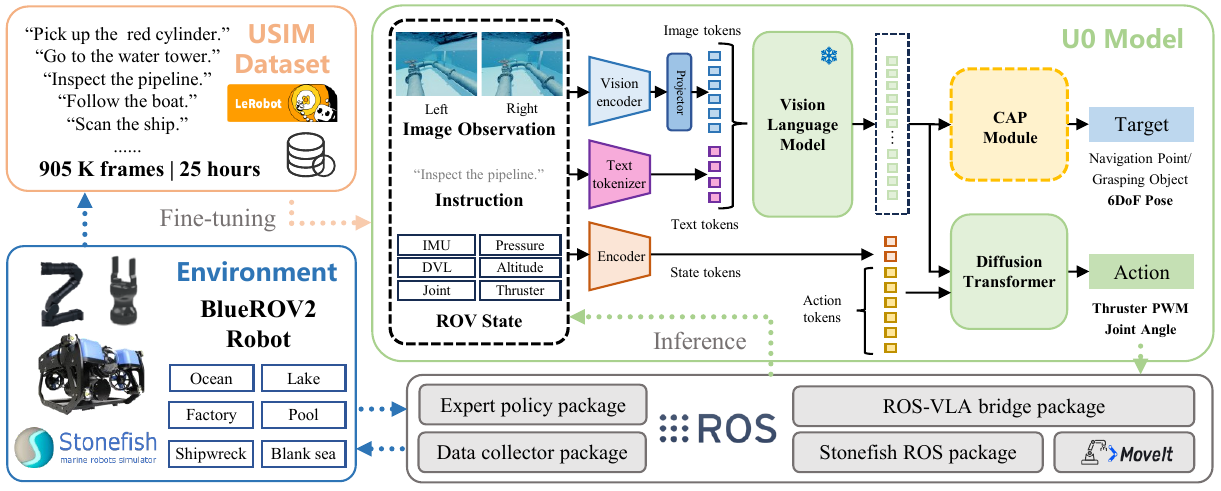}
   \caption{Overall Framework. Diverse underwater scenarios and a BlueROV2 robot equipped with a manipulator are first constructed within the Stonefish simulator. Data collection and control are managed via ROS, yielding the USIM dataset of 905K frames (approximately 25 hours) across 20 tasks. The U0 model is then designed, which integrates a DiT action module and a CAP module, generating concurrent target perception and robot action outputs.
   }
   \label{fig:framework} 
\end{figure*}

\section{Related Work}\label{sec:related_work}

\subsection{Data-Driven Underwater Autonomy}
Data-driven methods such as reinforcement learning (RL) and imitation learning (IL) have achieved significant progress in the underwater domain, covering path planning~\cite{yang_intelligent_2023}, multi-agent formation and obstacle avoidance~\cite{fang_multi-agent_2025}, target tracking~\cite{wang_target_2022, masmitja_dynamic_2023}, autonomous navigation~\cite{lin_uivnav_2024}, and mobile manipulation~\cite{gao_improved_2024, liu_self-improving_2025}. However, most of these approaches are designed as expert models--highly specialized systems utilizing task-specific observation and action spaces. This specialization results in a lack of versatility when facing diverse natural language instructions and cross-task scenarios. While large language models~\cite{buchholz_distributed_2025} and vision-based deep learning methods~\cite{gomez_chavez_underwater_2021} have been introduced to improve high-level cognition and human-robot interaction, they often lack adaptive mapping to low-level controllers. This fragmented technical trajectory limits the evolution of underwater robots toward high-level embodied intelligence. To bridge this gap, we integrate navigation and manipulation tasks into a unified state-action space, constructing U0 as a cohesive framework capable of understanding natural language and executing multiple tasks simultaneously.

\subsection{Vision-Language-Action Models}
Vision-language-action models have demonstrated exceptional generalization capabilities in terrestrial robotics by combining large-scale pre-training with embodied tasks. Representative models such as RT-2~\cite{brohan_rt-2_2023}, OpenVLA~\cite{kim_openvla_2024}, the $\pi$ series~\cite{black_0_2024, intelligence__05_2025}, and GR00T~\cite{nvidia_gr00t_2025} have proven that through end-to-end policy learning--utilizing architectures like vision-language models, mixture-of-experts, or dual systems--robots can achieve zero-shot generalization across diverse tasks. Recent studies have also extended VLA to unmanned aerial vehicle navigation by incorporating depth information and multi-stage training~\cite{sun_autofly_2026, wu_vla-_2025}. However, VLA models specifically designed for underwater robotics remain exceedingly rare. The primary challenge lies in the significant perceptual domain shift caused by underwater conditions~\cite{xu_nautilus_2025}. Addressing this, we introduce the CAP module in the training of U0, which incorporates target pose estimation as an auxiliary task, effectively enhancing the model's spatial perception and its adaptability to aquatic environments.

\subsection{Underwater Datasets}
Current underwater datasets primarily focus on perception-only or isolated navigation tasks. For instance, datasets like Caves~\cite{mallios_underwater_2017}, AQUALOC~\cite{ferrera_aqualoc_2019}, and AquaticVision~\cite{peng_aquaticvision_2025} support simultaneous localization and mapping (SLAM), while UIEB~\cite{li_underwater_2020} and the Marine Debris Dataset~\cite{singh_marine_2021} target image enhancement and segmentation. To mitigate data scarcity, synthetic alternatives like VAROS~\cite{georg_olofsson_zwilgmeyer_varos_2021} and MIMIR-UW~\cite{alvarez-tunon_mimir-uw_2023} have also been introduced. Recently, multi-task datasets have emerged, including MOUD~\cite{chu_multimodal_2025} (detection, segmentation, 3D reconstruction) and SubPipe~\cite{alvarez-tunon_subpipe_2024} (SLAM and inspection). As categorized in Table~\ref{dataset_comparison}, while these pioneering datasets establish a solid foundation, scaled multi-task data encompassing both manipulation and diverse skills—akin to terrestrial counterparts like Open X-Embodiment~\cite{collaboration_open_2025}—remains a valuable next step for full underwater autonomy. Given the prohibitive costs of real-world underwater data collection, we offer a practical alternative by developing an automated data synthesis pipeline. This pipeline generates USIM, a simulation-based multi-task dataset comprising 20 distinct tasks and 2,275 trajectories, thereby enhancing the adaptability and versatility of VLA models in dynamic aquatic environments.

\section{Method}\label{sec:method}

The overall framework of this work is illustrated in Fig.~\ref{fig:framework}, comprising an integrated underwater simulation environment, a multi-task dataset USIM, and the VLA model U0. Detailed descriptions of the task formulation, dataset construction, and model architecture are provided in the following.

\subsection{Task Formulation}
We aim to enable a single, unified policy $\pi$ to execute a diverse array of underwater missions, including navigation, grasping, transport, and tracking. The policy is formulated as $\pi: (V_t, L, S_t, A_t) \rightarrow A_{t+1}$, where $V_t$ denotes multi-view visual observations from two RGB cameras, $L$ is the natural language instruction, and $S_t$ represents the sensory data. This sensory input integrates inertial measurement unit (IMU), doppler velocity log (DVL), and pressure-based depth readings, as well as thruster commands and robotic arm joint states (angles and angular velocities). The action space $A_t$ and $A_{t+1}$ encompasses both desired propeller thrust intensities and robotic arm joint positions.

\subsection{USIM Dataset Construction}

Given the prohibitive costs associated with real-world underwater experiments, simulation offers an efficient alternative for data collection. Consequently, we developed a simulation-based pipeline to construct the USIM dataset using an automated collection process integrated with a stochastic exploration strategy. Specifically, we utilized the Stonefish simulator~\cite{grimaldi_stonefish_2025} to construct 9 distinct scenarios, including seabed environment, subsea pipeline, industrial pool, solar charging station, lake environment, open sea surface, underwater factory, modern shipwreck, and ancient shipwreck. Furthermore, a BlueROV2 platform, optionally equipped with a 4-degree-of-freedom (DoF) arm and a parallel gripper, was deployed. Leveraging the Stonefish ROS package~\cite{cieslak_stonefish_2019}, we established a standardized framework that remains compatible with other underwater simulators supporting ROS protocols.

\begin{figure}
   \centering
   \includegraphics[width=\columnwidth]{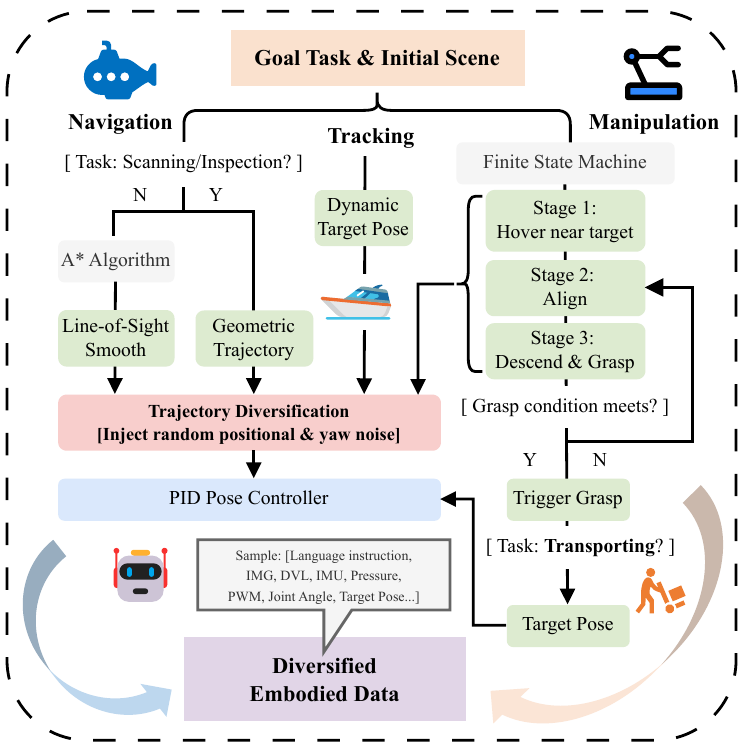}
   \caption{Automated data collection pipeline. Hand-crafted expert policies generate diverse navigation and manipulation trajectories in simulation. Stochastic noise injection is integrated to automatically capture sub-optimal behaviors and recovery sequences.}
   \label{fig:strategy}
\end{figure}

To ensure data diversity, we implemented scene randomization and exploratory collection strategies. Specifically, scene randomization modules were utilized to vary object placement and lighting conditions. For data collection, we employed hand-crafted expert strategies that leverage ground-truth state information available only in simulation, as shown in Fig.~\ref{fig:strategy}. These strategies enables fully automated data collection and, by introducing stochastic exploration, ensures that the dataset encompasses diverse trajectories, including sub-optimal states and recovery behaviors, which are critical for training robust policies.

Under this framework, we constructed the USIM dataset, comprising 20 tasks across 9 scenarios, with a total of 905K frames and 2275 trajectories (sampled at 10 Hz, approximately 25 hours). The 20 tasks include 12 grasping, 6 navigation, 1 dynamic tracking, and 1 transport task. Table~\ref{dataset_task_details} provides a detailed breakdown of the task distributions. We intentionally weighted the dataset toward grasping tasks due to the inherent complexity of 3D mobile manipulation. Since trajectories of mobile manipulation involve significant robot locomotion, they concurrently serve to improve the model's fundamental motion control, thereby benefiting navigation performance. Table~\ref{dataset_comparison} highlights the unique advantages of USIM, specifically its unprecedented scale, diverse task coverage, and the seamless integration of language descriptors with precise action labels, which are absent in previous underwater datasets.

\subsection{U0 Model Architecture}
U0 is a general-purpose underwater VLA model initialized with the pre-trained weights of Isaac-GR00T N1.5~\cite{nvidia_gr00t_2025}, as illustrated in Fig.~\ref{fig:framework}. The model accepts multi-modal data as input, where visual images and language instructions are processed through a VLM and mapped to 2048-dimensional embedding. Simultaneously, additional sensor modalities are encoded via a multi-layer perceptron (MLP), while robot action data are mapped to features of the same dimension through an MLP following the addition of positional embeddings. These features are then fed into a diffusion transformer (DiT) action module to generate action chunks. Concurrently, the VLM features are passed through the CAP module. By leveraging convolution-attention mechanisms, this module enhances visual feature representations to improve the model's ability to correlate underwater motion states with task objectives. Details regarding multi-sensor integration, the CAP module, and training objectives are elaborated below.

\begin{table}
    \centering
    \caption{Data Distribution of USIM}
    \label{dataset_task_details}
    \renewcommand\arraystretch{1.1}
    \begin{tabular}{lccc}
        \toprule
        Category & Total frames & Episodes & Average duration \\ 
        \midrule
        Grasping & 585K (64.6\%) & 1560 & 37.5 $\pm$ 8.7~s \\ 
        Navigation & 231K (25.5\%) & 520 & 44.4 $\pm$ 24.3~s \\ 
        Transport & 66K (7.3\%) & 130 & 51.1 $\pm$ 5.0~s \\ 
        Tracking & 23K (2.5\%) & 65 & 35.5 $\pm$ 2.5~s \\
        \midrule
        Total & 905K (100.0\%) & 2275 & 39.8 $\pm$ 14.3~s \\
        \bottomrule
    \end{tabular}
\end{table}

\subsubsection{Multi-Source Sensing and Unified Action Space}
Unlike terrestrial robots, underwater robots operate predominantly in a floating state and must coordinate locomotion with manipulator actions. They also frequently execute depth changes, in contrast to most ground robots limited to horizontal planes. To address these challenges, we incorporate underwater-specific sensors into the state space for accurate localization, including depth sensor, IMU, and DVL. Furthermore, dual-view camera images are provided as input to enhance visual coverage. The action space of underwater robots also differs significantly from terrestrial counterparts. Beyond manipulator control, locomotion is primarily achieved through multiple thrusters, whose control signals occupy a distinct force space. We normalize the thruster PWM signals and concatenate them with the manipulator joint angles to construct a unified action space. Both thruster commands and joint states are included in the state space to improve the model's temporal understanding of the robot's current physical configuration.

\subsubsection{Convolution-Attention-based Perception Module}

\begin{figure}
   \centering
   \includegraphics[width=\columnwidth]{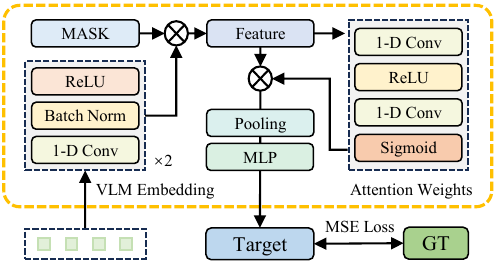}
   \caption{The structure of the CAP module. Through combining convolutional layers and attention pooling, feature focus is refined for enhanced perception.}
   \label{fig:module}
\end{figure}

Due to the vast spatial scales and sparse visual features in aquatic environments, we designed the CAP module to bolster visual comprehension and spatial awareness. Guided by VLM features, this module is optimized during training to predict a 6-DoF pose. Specifically, for grasping tasks, this represents the pose of the target objects; for navigation, it represents the next waypoint; for transport tasks, it predicts the object pose during grasping and the target waypoint during transfer; and for tracking, it predicts the dynamic pose of the tracked target. By explicitly supervising these points of interest, we direct the model's attention toward task-relevant features. The architecture of the CAP module, illustrated in Fig.~\ref{fig:module}, consists of convolutional layers, convolution-based attention weighting, pooling operations, and a multilayer perceptron (MLP). The computation process is formulated as
\begin{equation}
    \begin{aligned}
        \phi_t = \text{VLM}\left( V_t, L \right), \quad
        F = \text{Conv}\left(\phi_t, \text{MASK}\right), \\
        \text{Att} = \text{Conv}(F), \quad
        T = \text{MLP}\left(\text{Pool}\left(F \cdot \text{Att}\right)\right),
    \end{aligned}
\end{equation}
where $\text{Conv}(\cdot)$ and $\text{Pool}(\cdot)$ refer to convolutional and pooling operations respectively, while $\text{MASK}$ prevents computation over padding features. $\text{Att}$ represents channel-wise attention, $F$ is the intermediate feature map, and $T$ is the predicted 6-DoF target.

\subsubsection{Ground Truth and Training Objectives}
Based on our task formulation, since the policy outputs a unified action space, we construct a standardized ground truth (GT) across all tasks to supervise training. For action module, We adopt an action-chunking approach where actions for $H$ consecutive timesteps are treated as the GT. Each action includes the thruster intensities and joint angles for that moment. Following the flow-matching objective, we sample a noise vector and a denoising timestep $\tau$ to construct the noisy action
\begin{equation}
    A_t^\tau = \tau A_t + (1 - \tau)\epsilon,
\end{equation}
where $\epsilon$ is the sampled noise and $A_t$ is the GT action. The velocity field corresponding to this noisy action is defined as $v_\tau = A_t - \epsilon$. The DiT action module $V_{\theta}$ is trained to predict this velocity field, resulting in the action loss
\begin{equation}
    L_{\text{action}} = \mathbb{E}_\tau \left[ \lVert V_{\theta}(\phi_t, A_t^\tau, S_t) - v_\tau \rVert^2 \right].
\end{equation}
During inference, a Gaussian noise sample is used as the initial action input, and the final action is obtained through $K$-step forward Euler integration
\begin{equation}
    A_t^{\tau+1/K} = A_t^\tau + \frac{1}{K} V_{\theta}(\phi_t, A_t^\tau, S_t).
\end{equation}
This requires the DiT to perform $K$ forward passes. In this work, we set $H=16$ and $K=4$.

For CAP module, target poses are treated as the GT. To enable effective reasoning relative to the robot's current state, we represent target poses in a robot-centric coordinate system. Let the 6-DoF poses of the target and the robot in the world coordinate system be
\begin{equation}
    p_g = (R_g, \mathbf{t}_g), \quad p_r = (R_r, \mathbf{t}_r),
\end{equation}
where $R \in SO(3)$ denotes the rotation matrix and $\mathbf{t} \in \mathbb{R}^3$ the translation vector. The target pose in the robot-centric frame is
\begin{equation}
    T_{\text{gt}} = \big(R_r^\top R_g, R_r^\top (\mathbf{t}_g - \mathbf{t}_r)\big).
\end{equation}
The CAP training loss is
\begin{equation}
    L_{\text{CAP}} = \mathbb{E} \left[ \lVert T - T_{\text{gt}} \rVert^2 \right].
\end{equation}
The total training objective is to minimize $L$, which is the weighted sum of the action and perception losses
\begin{equation}
    L = L_{\text{action}} + \alpha L_{\text{CAP}},
\end{equation}
where the weighting factor $\alpha$ is set to $1.0$.

\section{Experiments}\label{sec:exp}
In this section, we define a comprehensive set of evaluation metrics and conduct experiments to assess the performance of the proposed framework across multiple tasks. To demonstrate the efficacy of our approach, we compare our framework against SOTA VLA methods and perform ablation studies to validate the contribution of each component.

\subsection{Data Split}
We allocated 77\% (697K frames) of USIM for training and 23\% (208K frames) for testing. To ensure evaluation integrity, both sets were collected under identical environmental conditions and follow the same statistical distribution.

\subsection{Implementation Details}
U0 was trained with a total batch size of 256 for 50000 steps on a server equipped with 8 NVIDIA H20 GPUs. In addition to U0, OpenVLA~\cite{kim_openvla_2024}, $\pi_{0.5}$~\cite{intelligence__05_2025}, and GR00T N1.5~\cite{nvidia_gr00t_2025} are selected as the comparison methods. These models were also fine-tuned on the USIM training set.

\subsection{Evaluation Metrics and Setup}
We evaluate model performance across both offline and online dimensions, adopting standard metrics widely established in vision-language-action and vision-language navigation tasks. Offline evaluation assesses policy imitation accuracy on the test dataset, primarily focusing on the action prediction error (APE), which is the mean squared error between the predicted and ground-truth actions. Online evaluation measures closed-loop control efficacy within the simulation environment using five complementary metrics: success rate (SR), stage success rate (SSR), success-weighted path length (SPL)~\cite{anderson_evaluation_2018}, average success duration (ASD), and mean target distance (MTD). For the deployment setup, evaluations are conducted on a server equipped with an Intel Xeon Silver 4214R CPU and four NVIDIA RTX 3090 GPUs. To ensure optimal performance, we customize the control and inference loops tailored to the architectural characteristics of each model. Specifically, for GR00T N1.5 and U0, the policy operates at a control frequency of 10~Hz, with model inference triggered every 1.6~s to predict 16-step action chunks. For $\pi_{0.5}$, the model performs recurrent inference at 10~Hz, with the robot executing only the first step of the predicted action chunk at each interval. Conversely, for OpenVLA, the policy operates in a closed-loop manner where the model executes inference as rapidly as possible and immediately applies the predicted action to the robot.

\subsection{Offline Evaluation and Comparison}

\begin{table}
    \centering
    \caption{Offline APE Results on USIM Test Set}
    \label{tab:offline_action_mse}
    \scriptsize
    \renewcommand{\arraystretch}{1.16}
    \setlength{\tabcolsep}{3.0pt}
    \begin{tabular}{llccccc}
        \toprule
        Model & Setting & Navigation & Grasping & Transport & Tracking & Overall \\
        \midrule
        \multirow{2}{*}{OpenVLA}
        & Zero-shot & 0.5672 & 0.1872 & 0.2160 & 0.3810 & 0.2920 \\
        & Fine-tuned & 0.3281 & 0.0768 & 0.0666 & 0.0827 & 0.1410 \\
        \midrule
        \multirow{2}{*}{$\pi_{0.5}$}
        & Zero-shot & 0.2363 & 0.1217 & 0.1181 & 0.0591 & 0.1496 \\
        & Fine-tuned & 0.2087 & 0.0423 & 0.0475 & 0.0528 & 0.0861 \\
        \midrule
        \multirow{2}{*}{GR00T N1.5}
        & Zero-shot & 0.3459 & 0.1220 & 0.1452 & 0.1906 & 0.1834 \\
        & Fine-tuned & 0.0974 & 0.0161 & 0.0158 & \textbf{0.0272} & 0.0374 \\
        \midrule
        U0 (Ours) & Fine-tuned & \textbf{0.0935} & \textbf{0.0154} & \textbf{0.0141} & 0.0310 & \textbf{0.0359} \\
        \bottomrule
    \end{tabular}
\end{table}

We compared the proposed model with several baselines across 525 trajectories in the USIM test set, with results summarized in Table~\ref{tab:offline_action_mse}. In the zero-shot setting, OpenVLA, $\pi_{0.5}$, and GR00T N1.5 all exhibit severe APE. This reveals a substantial perceptual and action-space domain shift between terrestrial and complex underwater environments, highlighting the difficulty of direct cross-domain transfer. However, after fine-tuning on our USIM dataset, the APE of all baseline models improves drastically, yielding an average error reduction of 59.7\% across the three competitive methods. This performance leap demonstrates the effectiveness and necessity of the USIM dataset. It successfully bridges the underwater perceptual barrier, empowering existing VLA models with cross-modal adaptability and policy generalization across diverse aquatic multi-task scenarios.

Among all the fine-tuned models, our proposed U0 model achieves the SOTA performance, reducing the overall APE to a minimum of 0.0359. Although GR00T N1.5 already exhibits exceptionally low error due to its robust design, U0 further pushes the performance boundary (0.0359 vs.\ 0.0374). This is attributed to the integration of the CAP module and auxiliary pose estimation tasks, which allow the model to capture sparse underwater features more accurately and enhance 3D spatial awareness.

On the whole, the offline evaluation results validate the necessity and feasibility of utilizing the USIM dataset to train underwater embodied policies, while fully demonstrating the superiority of the CAP-based auxiliary training strategy in improving high-precision aquatic control.

\subsection{Closed-Loop Autonomy and Efficiency Analysis}

\begin{table*}
    \centering
    \caption{Online Evaluation Performance Comparison}
    \label{tab:online_cmp}
    \scriptsize
    \renewcommand{\arraystretch}{1.16}
    \setlength{\tabcolsep}{3.0pt}
    \resizebox{0.8\textwidth}{!}{%
    \begin{tabular}{lccccccccc}
        \toprule
        \multirow{2}{*}{Model}
        & \multicolumn{2}{c}{Navigation}
        & \multicolumn{2}{c}{Grasping}
        & \multicolumn{2}{c}{Transporting}
        & \multicolumn{2}{c}{Tracking}
        & \multicolumn{1}{c}{Overall SR} \\
        \cmidrule(lr){2-3} \cmidrule(lr){4-5} \cmidrule(lr){6-7} \cmidrule(lr){8-9}
        & SR$\uparrow$ & SPL$\uparrow$ & SR$\uparrow$ & ASD$\downarrow$ & SR$\uparrow$ & SSR$\uparrow$ & SR$\uparrow$ & MTD$\downarrow$ & (succeed/total trials) \\
        \midrule

        OpenVLA
        & 25.6\% & $0.70 \pm 0.21$
        & 0.0\% & --
        & 0.0\% & 0.0\%
        & 5.0\% & $4.02 \pm 0.00$\,\text{m}
        & 6.0\% (42/700) \\

        $\pi_{0.5}$
        & 46.9\% & \textbf{0.87 $\pm$ 0.18}
        & \textbf{37.1\%} & 97.3\,s
        & 20.0\% & $45.0\%$
        & 10.0\% & $4.52 \pm 0.06$\,\text{m}
        & 37.6\% (263/700) \\

        GR00T N1.5
        & 76.3\% & $0.69 \pm 0.21$
        & 24.0\% & 92.3\,s
        & 15.0\% & 30.0\%
        & 30.0\% & $4.16 \pm 0.14$\,\text{m}
        & 35.6\% (249/700) \\

        U0 (Ours)
        & \textbf{87.5\%} & $0.71 \pm 0.23$
        & 30.0\% & \textbf{87.6\,s}
        & \textbf{25.0\%} & \textbf{45.0\%}
        & \textbf{40.0\%} & \textbf{3.61 $\pm$ 0.35}\,\text{m}
        & \textbf{43.1\% (302/700)} \\

        \bottomrule
    \end{tabular}}
\end{table*}

\begin{figure}
   \centering
   \includegraphics[width=\columnwidth]{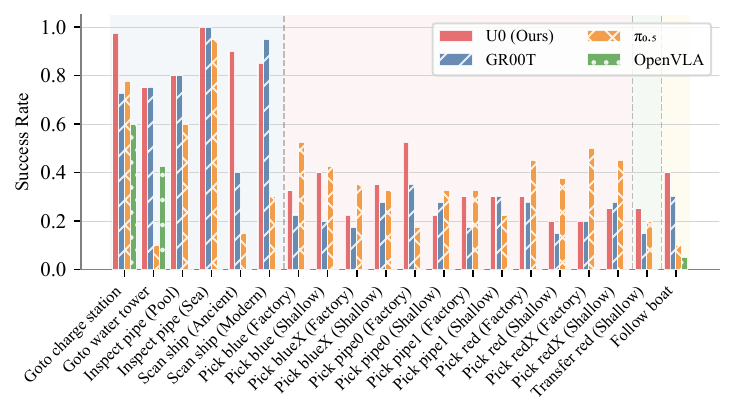}
   \caption{Breakdown of success rates across 20 individual tasks.}
   \label{fig:task_details}
\end{figure}

\begin{figure*}
   \centering
   \includegraphics[width=\textwidth]{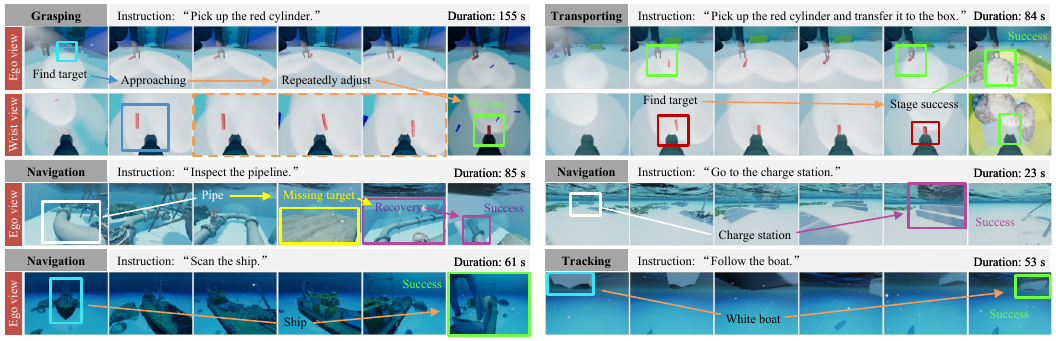}
   \caption{Real-time execution traces of U0 across various tasks.}
   \label{fig:online_traces}
\end{figure*}

We conducted closed-loop evaluations in the simulation environment. As the first VLA framework designed for multi-task underwater operations, we utilize U0 to establish the initial online performance benchmark. Specifically, we conducted 40 independent trials for each task, except for the 4 inspection and 1 tracking tasks, which involved 20 trials each. Generous maximum execution times, ranging from 180 to 450~s, were tailored to each mission based on its difficulty, with any exceeding duration resulting in task failure. The positions of the robot, environmental obstacles, target objects, and waypoints were randomized to ensure statistical significance before each trial. Physically grounded success criteria were also defined. Grasping requires the robot to maintain a stable grip for at least 3s after relocation; navigation and transport success is defined by the robot or target object reaching a task-specific radius of the goal; and tracking success requires maintaining the dynamic target within the field of view and within a safety distance threshold for a required cumulative duration.

The success rates across 20 distinct tasks, compiled from a total of 700 independent trials, are summarized in Table~\ref{tab:online_cmp}. As observed, fine-tuning on the USIM dataset successfully unlocks the underwater execution capabilities of all three terrestrial VLA baselines, underscoring the dataset's efficacy. Among the baselines, $\pi_{0.5}$ exhibits competitive performance in grasping-related tasks, while GR00T N1.5 excels in navigation missions. Conversely, OpenVLA underperforms across the board; this is primarily attributed to its high computational overhead and lower inference frequency, which incur noticeable control latency during complex 3D locomotion and fine-grained manipulation.

Benefiting from the integration of the CAP module, our proposed U0 achieves a SOTA overall SR of 43.1\%, outperforming the strongest baseline by a margin of 5.5\%. Fig.~\ref{fig:task_details} breaks down the performance across all 20 individual tasks. Specifically, the model demonstrates exceptional proficiency in navigation tasks, achieving an 87.5\% SR and an average SPL of 0.71. This high SPL indicates that the executed trajectories closely approximate the optimal geometric paths, reflecting superior path efficiency. In manipulation tasks, U0 also delivers robust performance, securing a 30.0\% SR in grasping. Furthermore, in the long-horizon transport task, which demands a seamless integration of navigation and manipulation capabilities, U0 achieves a 25.0\% SR—surpassing comparison methods by 5.0\%. In dynamic target tracking, U0 boosts the SR by 10.0\% and maintains a significantly tighter tracking loop, reducing the MTD to 3.61~m.

\subsection{Case Studies and Robustness Verification}

Fig.~\ref{fig:online_traces} provides snapshots of the robot's performance during real-time execution. For grasping and transport tasks, the figures display two distinct visual fields: the ego view and the wrist view. Conversely, in navigation and tracking tasks, where the robot is not equipped with a manipulator, the dual views are natively provided by a stereo camera setup on the robot body; thus, only one of these ego views is displayed.

The grasping sequences reveal that the dynamic nature of the robot—requiring simultaneous station-keeping and manipulation—imposes rigorous demands on the model's motion control. Notably, the trajectories demonstrate that U0 can autonomously initiate re-grasping attempts following an initial failure, signifying the emergence of self-recovery behaviors. The average duration for a successful grasp during online testing is 87.6~s, which, though the fastest among all evaluated models, is still over twice that of the expert strategy (37.5~s), highlighting potential for further policy optimization.

\subsection{Ablation Studies}

\newcommand{\TaskCategoryFull}[1]{%
    \multicolumn{2}{@{}l@{}}{\scriptsize\textbf{#1}}\\[0.2mm]
}

\newcommand{\SharedHeaderPair}[4]{%
    \multicolumn{2}{@{}c@{}}{%
        \begin{tabular*}{\linewidth}{
            @{}
            p{0.466\linewidth}
            @{\hspace{0.55mm}}
            !{\vrule width 0.25pt}
            @{\hspace{0.55mm}}
            p{0.466\linewidth}
            @{}
        }
            \begin{minipage}[t]{\linewidth}
            \scriptsize
            \setlength{\tabcolsep}{1.2pt}
            \setlength{\aboverulesep}{0.6pt} 
            \setlength{\belowrulesep}{0.6pt} 
            \begin{tabular*}{\linewidth}{
                @{\hspace{2mm}}
                l
                @{\extracolsep{\fill}}
                c
                c
                @{\hspace{2mm}}
            }
                \toprule
                \textbf{Model} & \textbf{#1} & \textbf{#2}
            \end{tabular*}
            \end{minipage}%
            &
            \begin{minipage}[t]{\linewidth}
            \scriptsize
            \setlength{\tabcolsep}{1.2pt}
            \setlength{\aboverulesep}{0.6pt}
            \setlength{\belowrulesep}{0.6pt}
            \begin{tabular*}{\linewidth}{
                @{\hspace{2mm}}
                l
                @{\extracolsep{\fill}}
                c
                c
                @{\hspace{2mm}}
            }
                \toprule
                \textbf{Model} & \textbf{#3} & \textbf{#4}
            \end{tabular*}
            \end{minipage}%
        \end{tabular*}%
    }\\[0.1mm]
}

\newcommand{\AblationDataBlock}[2]{%
\begin{minipage}[t]{\linewidth}
\centering
\scriptsize
\setlength{\tabcolsep}{1.2pt}
\renewcommand{\arraystretch}{0.92} 
\setlength{\aboverulesep}{0.6pt}  
\setlength{\belowrulesep}{0.6pt}
\begin{tabular*}{\linewidth}{
    @{\hspace{2mm}}
    l
    @{\extracolsep{\fill}}
    c
    c
    @{\hspace{2mm}}
}
    \toprule
    \multicolumn{3}{@{\hspace{1.2mm}}l@{\hspace{1.2mm}}}{\textit{#1}} \\
    \midrule
    #2
\end{tabular*}
\end{minipage}%
}

\newcommand{\AblationDataBlockLast}[2]{%
\begin{minipage}[t]{\linewidth}
\centering
\scriptsize
\setlength{\tabcolsep}{1.2pt}
\renewcommand{\arraystretch}{0.92} 
\setlength{\aboverulesep}{0.6pt}
\setlength{\belowrulesep}{0.6pt}
\begin{tabular*}{\linewidth}{
    @{\hspace{2mm}}
    l
    @{\extracolsep{\fill}}
    c
    c
    @{\hspace{2mm}}
}
    \toprule
    \multicolumn{3}{@{\hspace{1.2mm}}l@{\hspace{1.2mm}}}{\textit{#1}} \\
    \midrule
    #2
    \bottomrule
\end{tabular*}
\end{minipage}%
}

\newcommand{\TaskCategoryPair}[2]{%
    \multicolumn{2}{@{}c@{}}{%
        \begin{tabular*}{\linewidth}{
            @{}
            p{0.466\linewidth}
            @{\hspace{1.35mm}}
            p{0.466\linewidth}
            @{}
        }
            {\scriptsize\textbf{#1}} & {\scriptsize\textbf{#2}}
        \end{tabular*}%
    }\\[0.2mm]
}

\begin{table}
    \centering
    \caption{Ablation Study on the CAP Module of U0}
    \label{tab:ablation_study}
    \scriptsize
    \setlength{\tabcolsep}{0pt}
    \renewcommand{\arraystretch}{1.0}

    \begin{tabular}{
        @{}
        p{0.466\linewidth}
        @{\hspace{0.55mm}}
        !{\vrule width 0.25pt}
        @{\hspace{0.55mm}}
        p{0.466\linewidth}
        @{}
    }

    \TaskCategoryFull{Navigation}
    \SharedHeaderPair{SR$\uparrow$}{SPL$\uparrow$}{SR$\uparrow$}{SPL$\uparrow$}

    \AblationDataBlock{Goto charge station}{
        w/  & \textbf{97.5\% (39/40)} & \textbf{0.611} \\
        w/o & 72.5\% (29/40) & 0.576 \\
    }
    &
    \AblationDataBlock{Goto water tower}{
        w/  & \textbf{75.0\% (30/40)} & 0.822 \\
        w/o & 75.0\% (30/40) & 0.859 \\
    }
    \\[0.1mm] 

    \AblationDataBlock{Inspect pipe (Pool)}{
        w/  & \textbf{80.0\% (16/20)} & 0.714 \\
        w/o & 80.0\% (16/20) & 0.789 \\
    }
    &
    \AblationDataBlock{Inspect pipe (Sea)}{
        w/  & \textbf{100.0\% (20/20)} & 0.547 \\
        w/o & 100.0\% (20/20) & 0.556 \\
    }
    \\[0.1mm]

    \AblationDataBlockLast{Scan ship (Ancient)}{
        w/  & \textbf{90.0\% (18/20)} & \textbf{0.978} \\
        w/o & 40.0\% (8/20)  & 0.760 \\
    }
    &
    \AblationDataBlockLast{Scan ship (Modern)}{
        w/  & 85.0\% (17/20) & 0.647 \\
        w/o & 95.0\% (19/20) & 0.658 \\
    }
    \\[0.2mm]

    \TaskCategoryFull{Grasping}
    \SharedHeaderPair{SR$\uparrow$}{ASD$\downarrow$}{SR$\uparrow$}{ASD$\downarrow$}

    \AblationDataBlock{Pick blue (Factory)}{
        w/  & \textbf{32.5\% (13/40)} & 113.35 \\
        w/o & 22.5\% (9/40)  & 83.13 \\
    }
    &
    \AblationDataBlock{Pick blue (Shallow)}{
        w/  & \textbf{40.0\% (16/40)} & 78.50 \\
        w/o & 20.0\% (8/40)  & 71.45 \\
    }
    \\[0.1mm]

    \AblationDataBlock{Pick blueX (Factory)}{
        w/  & \textbf{22.5\% (9/40)}  & \textbf{78.79} \\
        w/o & 17.5\% (7/40)  & 109.80 \\
    }
    &
    \AblationDataBlock{Pick blueX (Shallow)}{
        w/  & \textbf{35.0\% (14/40)} & 101.89 \\
        w/o & 27.5\% (11/40) & 63.08 \\
    }
    \\[0.1mm]

    \AblationDataBlock{Pick pipe0 (Factory)}{
        w/  & \textbf{52.5\% (21/40)} & \textbf{85.66} \\
        w/o & 35.0\% (14/40) & 97.23 \\
    }
    &
    \AblationDataBlock{Pick pipe0 (Shallow)}{
        w/  & 22.5\% (9/40)  & \textbf{79.53} \\
        w/o & 27.5\% (11/40) & 103.89 \\
    }
    \\[0.1mm]

    \AblationDataBlock{Pick pipe1 (Factory)}{
        w/  & \textbf{30.0\% (12/40)} & \textbf{90.88} \\
        w/o & 17.5\% (7/40)  & 100.38 \\
    }
    &
    \AblationDataBlock{Pick pipe1 (Shallow)}{
        w/  & \textbf{30.0\% (12/40)} & \textbf{86.56} \\
        w/o & 30.0\% (12/40) & 89.81 \\
    }
    \\[0.1mm]

    \AblationDataBlock{Pick red (Factory)}{
        w/  & \textbf{30.0\% (12/40)} & \textbf{91.73} \\
        w/o & 27.5\% (11/40) & 111.53 \\
    }
    &
    \AblationDataBlock{Pick red (Shallow)}{
        w/  & \textbf{20.0\% (8/40)}  & \textbf{48.81} \\
        w/o & 15.0\% (6/40)  & 90.54 \\
    }
    \\[0.1mm]

    \AblationDataBlockLast{Pick redX (Factory)}{
        w/  & \textbf{20.0\% (8/40)}  & 75.55 \\
        w/o & 20.0\% (8/40)  & 73.80 \\
    }
    &
    \AblationDataBlockLast{Pick redX (Shallow)}{
        w/  & 25.0\% (10/40) & \textbf{100.92} \\
        w/o & 27.5\% (11/40) & 107.64 \\
    }
    \\[0.2mm]

    \TaskCategoryPair{Transporting}{Tracking}
    \SharedHeaderPair{SR$\uparrow$}{SSR$\uparrow$}{SR$\uparrow$}{MTD$\downarrow$}

    \AblationDataBlockLast{Transfer red (Shallow)}{
        w/  & \textbf{25.0\% (10/40)} & \textbf{45.0\%} \\
        w/o & 15.0\% (6/40)  & 30.0\% \\
    }
    &
    \AblationDataBlockLast{Follow boat}{
        w/  & \textbf{40.0\% (8/20)} & \textbf{3.61} \\
        w/o & 30.0\% (6/20) & 4.16 \\
    }
    \\[0.2mm]

    \multicolumn{2}{@{}c@{}}{\scriptsize\textbf{Overall Success Rate (succeed/total)}}\\[0.1mm]
    \multicolumn{2}{@{}l@{}}{%
        \begin{minipage}[t]{\dimexpr 0.932\linewidth + 1.1mm + 0.25pt\relax}
        \centering
        \scriptsize
        \setlength{\tabcolsep}{1.2pt}
        \renewcommand{\arraystretch}{0.92}
        \setlength{\aboverulesep}{0.6pt}
        \setlength{\belowrulesep}{0.6pt}
        \begin{tabular*}{\linewidth}{
            @{\hspace{2mm}}
            l
            @{\extracolsep{\fill}}
            l
            @{\hspace{2mm}}
        }
            \toprule
            w/  & \textbf{43.1\% (302/700)} \\
            w/o & 35.6\% (249/700) \\
            \bottomrule
        \end{tabular*}
        \end{minipage}%
    }\\[0.2mm]

    \end{tabular}
\end{table}

To further validate the efficacy of our framework, we conducted extensive ablation studies. The variant trained without the CAP module are analyzed using the same protocol of 700 trials. Table~\ref{tab:ablation_study} presents detailed ablation results on 20 tasks. The data indicates that the CAP-based auxiliary training branch provides substantial performance gains across all task types, improving the overall success rate by 7.5\%. This suggests that providing explicit guidance regarding the model's points of interest during training significantly enhances the model's task comprehension.

\section{Conclusions and Future Work}\label{sec:conclusion}
In this paper, we propose a unified VLA framework to enhance the multi-task autonomy of underwater robots. At first, we establish USIM, a large-scale simulation-based dataset generated through an automated synthesis pipeline, which provides a diverse foundation for robust policy learning in dynamic aquatic environments. Widespread fine-tuning experiments demonstrated that USIM successfully unlocks the zero-to-one capability of existing terrestrial VLA models in executing underwater tasks, underscores its profound efficacy and cross-domain value. Building upon this dataset, we develop U0, a 3B-parameter VLA model specifically optimized for the unique embodiment and perception challenges of underwater robots. By incorporating a CAP module, U0 achieves enhanced spatial awareness, effectively bridging the gap between high-level instructions and low-level 6-DoF actuation. Our extensive closed-loop evaluations demonstrate that U0 significantly outperforms existing terrestrial-based VLA models, improving the overall success rate by 5.5\% and achieving an exceptional 87.5\% success rate in navigation, while exhibiting superior efficiency and robustness in long-horizon tasks and dynamic environments.

Future work will focus on incorporating supplementary modalities, such as sonar, to improve perception in low-visibility environments, eventually moving toward real-world deployment and field validation.










\bibliographystyle{IEEEtran}
\bibliography{reference/underwater_intelligence,reference/underwater_simulator,reference/vla_reference,reference/underwater_dataset}

\end{document}